\documentclass[11pt]{article}

\usepackage[preprint]{acl}

\usepackage{times}
\usepackage{latexsym}
\usepackage[T1]{fontenc}
\usepackage[utf8]{inputenc}
\usepackage{microtype}
\usepackage{inconsolata}
\usepackage{graphicx}

\usepackage{booktabs}
\usepackage{multirow}
\usepackage{amsmath}
\usepackage{amssymb}
\usepackage{enumitem}
\usepackage{algorithm}
\usepackage{algpseudocode}
\usepackage[most]{tcolorbox}
\usepackage{xcolor}
\usepackage{colortbl}

\definecolor{msBlue}{HTML}{1F5AA6}
\definecolor{msBlueLight}{HTML}{EAF1FB}
\definecolor{msGoldLight}{HTML}{FFF3D6}
\definecolor{msGreenLight}{HTML}{EAF6EF}
\definecolor{msRule}{HTML}{6F8199}
\definecolor{msMuted}{HTML}{6B7280}
\definecolor{msAlgBack}{HTML}{F7FAFE}
\definecolor{msAlgPhase}{HTML}{DCEBFA}

\makeatletter
\renewcommand\fs@ruled{%
  \def\@fs@cfont{\bfseries\color{msBlue}}%
  \let\@fs@capt\floatc@ruled
  \def\@fs@pre{{\color{msRule}\hrule height 0.7pt}\kern2pt}%
  \def\@fs@post{\kern2pt{\color{msRule}\hrule height 0.7pt}\relax}%
  \def\@fs@mid{\kern2pt{\color{msRule}\hrule height 0.45pt}\kern2pt}%
  \let\@fs@iftopcapt\iftrue}
\makeatother
\floatstyle{ruled}
\restylefloat{algorithm}
\algrenewcommand\algorithmicrequire{\textcolor{msBlue}{\textbf{Input:}}}
\algrenewcommand\algorithmicensure{\textcolor{msBlue}{\textbf{Output:}}}
\algrenewcommand\algorithmiccomment[1]{\hfill\footnotesize\textcolor{msMuted}{$\triangleright$~#1}}
\algnewcommand\algorithmicphase{}
\newcommand{\Phase}[1]{\Statex\smallskip\colorbox{msAlgPhase}{\strut\textcolor{msBlue}{\textit{#1}}}}

\newcommand{\blfootnote}[1]{%
  \begingroup
  \renewcommand\thefootnote{}\footnote{#1}%
  \addtocounter{footnote}{-1}%
  \endgroup
}

\title{MetaSkill-Evolve: Recursive Self-Improvement of LLM Agents\\
via Two-Timescale Meta-Skill Evolution}

\author{
 \textbf{Zefeng Wang\textsuperscript{*,1}},
 \textbf{Minxi Yan\textsuperscript{*,2}},
 \textbf{Jinhe Bi\textsuperscript{1}},
 \textbf{Sikuan Yan\textsuperscript{1}},
 \textbf{Volker Tresp\textsuperscript{1}},
 \textbf{Yunpu Ma\textsuperscript{1,3,4}}
\\
 \textsuperscript{1}LMU Munich,
 \textsuperscript{2}The Chinese University of Hong Kong,
 \textsuperscript{3}MCML,
 \textsuperscript{4}MemAgents Lab
\\
}

\begin{document}
\maketitle
\blfootnote{\textsuperscript{*}Equal contribution.}

\begin{figure*}[t]
\centering
\includegraphics[width=0.92\textwidth]{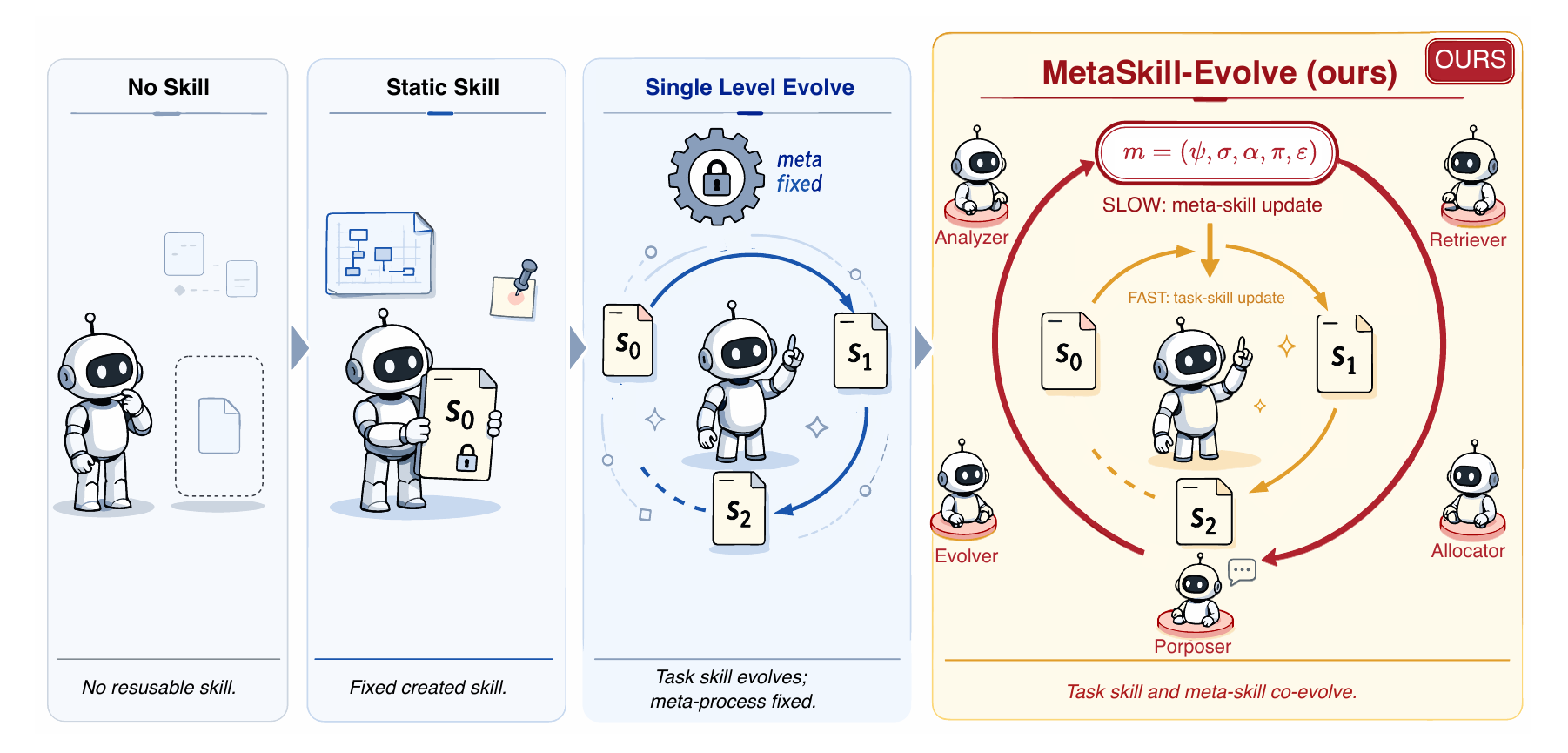}
\caption{\textbf{Four regimes of agent skill improvement.}
\emph{No-Skill}: no reusable skill memory. \emph{Static Skill}: a hand-authored $s_0$ held fixed (padlocked). \emph{Single Level Evolve}: the task skill evolves $s_0\!\to\!s_1\!\to\!s_2$, but the driving meta-process stays padlocked. \emph{MetaSkill-Evolve} (ours): a branch-level meta-skill $m=(\psi,\sigma,\alpha,\pi,\varepsilon)$ co-evolves on a slower outer ring via the \emph{same} five-agent pipeline that rewrites $s$, with no extra model and no extra framework.}
\label{fig:teaser}
\end{figure*}

\begin{abstract}
Recent LLM agents tackle increasingly long-horizon, open-ended tasks, and external skills, reusable procedural knowledge supplied to the agent, further extend this capability.  However, a fixed, hand-authored skill is rarely optimal, and cannot adapt to the diversity of tasks an agent encounters.  Self-improving agents address this by rewriting their own skill files from execution traces, yielding meaningful gains on challenging benchmarks.  Yet such self-evolution remains \emph{non-recursive}: it improves only the task skill (\emph{what} the agent does) while the improvement procedure (\emph{how} it improves) is authored once and held fixed.  We introduce \textbf{MetaSkill-Evolve}, a two-timescale framework that makes agentic skill improvement \emph{recursive}: every branch carries both a task skill $s$ and a branch-local meta-skill $m=(\psi,\sigma,\alpha,\pi,\varepsilon)$ whose five components parameterise the Analyzer, Retriever, Allocator, Proposer, and Evolver agents of the improvement pipeline.  Task skills evolve on a fast loop while the meta-skill evolves on a slower one under the \emph{same} pipeline applied to itself, with no additional model or objective.  With all five pipeline agents sharing a single frozen backbone, MetaSkill-Evolve outperforms no-skill, static-skill, and single-level evolution baselines on three agentic benchmarks (OfficeQA, SealQA, ALFWorld), improving held-out test accuracy over the raw backbone by +23.54, +16.09, and +1.92 points respectively.
\end{abstract}

\section{Introduction}
\label{sec:introduction}

Language model agents now tackle increasingly long-horizon, open-ended tasks, from document understanding and multi-step reasoning to tool use, yet they rarely succeed out of the box~\citep{yao2023reactsynergizingreasoningacting}.  A productive remedy is to equip the agent with a \emph{skill}: a curated, editable Markdown specification of reusable procedures, now a portable file-system artifact in widely deployed agent harnesses~\citep{wang2023voyager,zheng2025skillweaverwebagentsselfimprove}.  But a fixed, hand-authored skill is rarely optimal, and cannot anticipate the diversity of tasks an agent encounters.  Self-improvement systems such as EvoSkill~\citep{alzubi2026evoskill}, GEPA~\citep{agrawal2026gepareflectivepromptevolution}, and SkillWeaver~\citep{zheng2025skillweaverwebagentsselfimprove} address this by closing the loop with an analyze--propose--evolve pipeline that rewrites the skill after each failure trace, so that iteration by iteration the skill grows more capable.

These systems, however, evolve only \emph{what} the agent does, not \emph{how it evolves}: the artifact under optimization changes while the operator that optimizes it stays fixed. In the vocabulary of self-improving machines~\citep{good1965ultraintelligent,schmidhuber2006goedelmachinesselfreferentialuniversal}, they are self-improving but stop short of being \emph{recursively} self-improving. The meta-level logic is hardcoded in advance and shared by every branch throughout the run: how failures are diagnosed, which edits are proposed, how much search effort is allocated, whether cross-branch experience is reused, and how an approved edit is applied to disk (Fig.~\ref{fig:teaser}, third panel).  A branch therefore cannot improve the way it diagnoses failures: it applies the same procedure to every error, whether a misread table or a faulty calculation, and when that procedure yields the wrong fix, nothing in the loop can revise it.

A closer look at this rigidity suggests that two quantities govern evolutionary skill search.  The first is the \emph{current skill utility} $U(s)$, the score of the present skill on a validation batch.  The second is the \emph{meta-productivity} $P(m \mid s)$, the rate at which a branch generates stronger descendants under its current improvement policy $m$.  These are not the same: a skill may score well today yet sit in a branch whose meta-level policy produces weak children, while a moderately-performing skill may reside in a branch whose policy reliably moves scores up, making it the more promising line to extend even though its present score lags behind.  Optimizing only $U(s)$ ignores the second quantity entirely, and we hypothesize that this omission is a primary reason fixed-meta evolution stalls once repeated failures share a diagnosis style the meta-process cannot revise. This motivates the central question of our work:

\begin{tcolorbox}[
  enhanced,
  colback=msBlueLight!55!white,
  colframe=msRule,
  boxrule=0.45pt,
  arc=2pt,
  left=8pt,right=8pt,top=6pt,bottom=6pt,
  borderline west={1.4pt}{0pt}{msBlue},
]
\small
\textcolor{msBlue}{\textbf{\textsc{Research Question}}}\par\smallskip
\emph{Can the improvement procedure itself be evolved as a first-class object alongside the task skills it produces, using the same agentic pipeline?}
\end{tcolorbox}

To this end, we introduce \textbf{MetaSkill-Evolve} (Fig.~\ref{fig:teaser}, rightmost panel), a two-timescale evolutionary framework that lifts the improvement procedure into a learnable object, yielding a practical, bounded form of recursive self-improvement in which the improvement operator is reflexively applied to itself. Every branch carries a state $b = (s, m, h)$: a task skill $s$, a branch-level meta-skill $m = (\psi, \sigma, \alpha, \pi, \varepsilon)$, and an iteration history $h$.  The task skill evolves at every iteration on the fast timescale; the meta-skill evolves every $H$ iterations on the slow timescale, driven by how much the branch's last $H$ descendants improved, i.e., a running measure of whether its current improvement policy is still productive.  The five components of $m$ jointly parameterise the improvement loop: $\psi$ diagnoses and tags a failure trace, $\sigma$ controls cross-branch retrieval, $\alpha$ sets the per-iteration child budget, $\pi$ turns a diagnosis into a concrete edit proposal, and $\varepsilon$ applies an approved proposal to the on-disk skill files and verifies that the result is coherent.

Crucially, this adds no architectural component: each component of $m$ is a Markdown skill file identical in format to a task skill, so the \emph{same} five-agent pipeline (Analyzer, Retriever, Allocator, Proposer, Evolver) that rewrites $s$ is applied recursively to refine $m$ itself, closing the self-improvement loop on the very operator that performs it.  The meta-skill in turn sharpens frontier selection beyond pure utility: each candidate parent is scored by $\eta_1 U(s) + \eta_2 P(m \mid s) + \eta_3 N(b)$, where $N(b)$ discounts branches already selected many times, steering the search toward branches that are at once productive and underexplored.

We evaluate the resulting system on three agentic benchmarks that stress complementary capabilities: \textbf{OfficeQA}~\citep{opsahlong2026officeqa}, \textbf{SealQA}~\citep{pham2026sealqaraisingbarreasoning}, and \textbf{ALFWorld}~\citep{shridhar2021alfworld}.  All five pipeline agents share a single frozen Gemma-4 31B~\citep{google2026gemma431b} backbone, so any improvement is attributable to the evolved skills and meta-skills rather than to added model capacity or training.  Against No-Skill, Static-Skill, and a Single-Level-Evolution baseline that ablates meta-skill updates, MetaSkill-Evolve improves held-out test accuracy by +23.54 / +16.09 / +1.92 points on OfficeQA / SealQA / ALFWorld over No-Skill, and by +6.38 / +8.05 / +1.92 points over Single-Level Evolution.  On the two QA benchmarks the progression No-Skill $\to$ Static $\to$ Single-Level $\to$ Ours is monotonic; on ALFWorld the backbone is already near ceiling on the held-out split, so the margins are small and the static skill is roughly neutral.

In summary, our contributions are:
\begin{enumerate}[leftmargin=*,itemsep=2pt,topsep=2pt]
    \item \textbf{Two-timescale framework} separating fast task-skill from slow meta-skill evolution, with meta-productivity $P(m\mid s)$ as the slow objective alongside task utility $U(s)$.  
    \item \textbf{Five-agent evolution pipeline} that extends the fixed analyze$\to$propose$\to$evolve loop with two typed stages they lack: a Retriever ($\sigma$) for cross-branch sharing and an Allocator ($\alpha$) for an adaptive per-parent child budget, giving Analyzer, Retriever, Allocator, Proposer, Evolver.
    \item \textbf{Recursive self-improvement via typed meta-skills.} $m{=}(\psi,\sigma,\alpha,\pi,\varepsilon)$ as meta-skills, evolved by the \emph{same} five-agent pipeline as task skills, a bounded, one-level recursion that needs no new model or objective.
    \item \textbf{Meta-aware frontier selection} that scores each candidate parent by utility $U(s)$, meta-productivity $P(m\mid s)$, and branch novelty $N(b)$, steering search toward branches that are at once productive and underexplored.
\end{enumerate}

\section{Related Work}
\label{sec:related}

\paragraph{Skill-based agents and skill self-improvement.}
A \emph{skill} is a reusable, named procedure that augments a frozen LLM agent and is now a portable artifact across agent harnesses \citep{wang2023voyager,zheng2025skillweaverwebagentsselfimprove,opsahlong2026officeqa,li2026skillsbenchbenchmarkingagentskills,jiang2026sok,zhao2026nl2codestructuredsurveymultimodal}.  A growing line closes the loop with execution feedback (reflection, distillation, analyze$\to$propose$\to$evolve and RL-driven rewrites, trajectory and memory-based methods) \citep{shinn2023reflexion,zhao2024expelllmagentsexperiential,alzubi2026evoskill,xia2026skillrlevolvingagentsrecursive,shi2026skill1,lin2025seagentselfevolutiontrajectoryoptimization,zhang2026memskilllearningevolvingmemory,fang2026mempexploringagentprocedural,ni2026trace2skill,tian2026skillscoach,si2026contextskillslanguagemodels,huang2025loongsynthesizelongchainofthoughts,yang2026alignsaeconceptalignedsparseautoencoders,ma-etal-2026-self,wan2025hyperion}, while a parallel sub-line grows a skill \emph{library} rather than a single artifact \citep{yang2026autoskill,Liu2026SkillForgeFD,shen2026skillfoundry,wang2026skillx,ma2026skillclaw,zhang2026coevoskillsselfevolvingagentskills}.  Throughout, the \emph{improvement procedure} (how failures are diagnosed, edits scoped, search allocated, experience reused) is authored once and held fixed. MetaSkill-Evolve makes that procedure a first-class, branch-local object, co-evolved by the same pipeline.

\paragraph{Recursive self-improvement.}
Improving a system's own capacity to improve traces to ultraintelligent machines~\citep{good1965ultraintelligent,wan2025magicwordssharpnessawareprompt,mid} and the G\"odel machine, which rewrites its own code under a proof of improvement~\citep{schmidhuber2006goedelmachinesselfreferentialuniversal}. LLM-era instances drop the proof, improving a code-improvement scaffold~\citep{zelikman2024selftaughtoptimizerstoprecursively}, co-evolving prompts and their mutation-prompts~\citep{fernando2023promptbreeder}, or rewriting their own designs~\citep{hu2025automateddesignagenticsystems,zhang2026darwingodelmachineopenended}. These systems recurse on \emph{code} or \emph{prompts} under one global policy.  MetaSkill-Evolve instead recurses on \emph{skill files}: the operator is a five-agent pipeline parameterised by a branch-local meta-skill that the same pipeline refines: a bounded, one-level recursion that adds no model and keeps a per-lineage rather than global policy.

\paragraph{Prompt, textual, and evolutionary optimisation.}
Another body of work treats the agent's prompt or context as the object of optimisation. Self-feedback, LLM optimisers, prompt compilation, textual gradients, evolutionary prompt search, and in-context search all improve the instructions seen by a fixed agent \citep{madaan2023selfrefine,Bi2025PRISMSI,yang2024largelanguagemodelsoptimizers,khattab2023dspy,pryzant2023automatic,zhou2023largelanguagemodelshumanlevel,yuksekgonul2024textgrad,guo2025evopromptconnectingllmsevolutionary,Bi2025CoTKineticsAT,agrawal2026gepareflectivepromptevolution,lee2025feedbackdescent,ouyang2026reasoningbankscalingagentselfevolving,ye2026metacontextengineeringagentic,bi-etal-2025-llava}.  Population-based program search makes the search object more executable by pairing behavioural archives with verifiers and recent extensions to coding agents and open-ended evolution \citep{holland2025alphaevolvecodingagent,lange2025shinkaevolveopenendedsampleefficientprogram,wang2025thetaevolvetesttimelearningopen,he2026evotestevolutionarytesttimelearning,bi2026echorl}.  All rewrite a single artifact under one fixed rule. We instead let the rule diverge across branches and evolve on its own timescale; our frontier score $\eta_1 U_v {+} \eta_2 \hat{P}_v {+} \eta_3 N_v$ uses $\hat{P}_v$ as a quality-diversity descriptor preserving improvement-policy diversity.

\section{MetaSkill-Evolve}
\label{sec:methods}

This section presents MetaSkill-Evolve. We first formalise task-skill evolution and its utility objective (\S\ref{sec:formulation}), then introduce the per-branch \emph{meta-skill} that parameterises the search (\S\ref{sec:meta-skill}). We next describe the persistent evolution graph and the score that decides which branch to expand (\S\ref{sec:graph}). The final two subsections detail the algorithm's two timescales: a \emph{fast loop} that evolves task skills on the selected parent (\S\ref{sec:fast}), and a \emph{slow loop} that, every $H$ iterations, evolves the meta-skill itself by reapplying the same five-agent pipeline to the hidden meta-skill files (\S\ref{sec:slow}).

\subsection{Problem Formulation}
\label{sec:formulation}

Let $\mathcal{T}$ be a task with inputs $x$ and expected outputs $y$.  A \emph{task skill} $s$ is a Markdown-format LLM-agent program specifying procedures, tools, and heuristics for $\mathcal{T}$.  Writing $A_s$ for the agent that executes skill $s$, its utility is the expected task reward \begin{equation}
  U(s) \;=\; \mathbb{E}_{(x,y)\sim\mathcal{T}}\!\left[\,r\big(A_s(x),\,y\big)\right],
\label{eq:utility}
\end{equation}
where $r(\cdot,\cdot)\in[0,1]$ scores a prediction against the reference output. Because $\mathcal{T}$ is accessible only through samples, $U(s)$ is estimated as accuracy on a held-out validation batch. 

\subsection{Branch State and Meta-Skill}
\label{sec:meta-skill}
Each task-skill iteration turns a failure into a skill edit through a fixed five-step procedure, i.e., diagnose, retrieve, allocate, propose, execute. MetaSkill-Evolve makes that procedure adaptive by attaching to each branch a \emph{meta-skill} $m$ that parameterises all five steps.  A \emph{branch state} is $b = (s, m, h)$, where $h$ is the branch's iteration history, and
\begin{equation}
  m = (\psi,\; \sigma,\; \alpha,\; \pi,\; \varepsilon).
\label{eq:meta}
\end{equation}
Each component is itself a Markdown-format LLM-agent program (a \texttt{SKILL.md} file) consumed by exactly one specialist agent: \begin{itemize}[leftmargin=*,topsep=2pt,itemsep=1pt,parsep=0pt]
\item $\psi$ -- \emph{diagnosis policy} (Analyzer): maps failures to a tag $\phi$ and free-form analysis $a$.
\item $\sigma$ -- \emph{sharing policy} (Retriever): selects same-branch and cross-branch inspirations matching $\phi$.
\item $\alpha$ -- \emph{allocation policy} (Allocator): sets the child budget $K \in [1, K_{\max}]$ per step.
\item $\pi$ -- \emph{edit-proposal policy} (Proposer): emits an edit $\delta$ conditioned on the worst case, analysis, and retrieved inspirations, i.e., $(f, a, \mathcal{I})$.
\item $\varepsilon$ -- \emph{edit-executor policy} (Evolver): writes $\delta$ to disk and verifies the result.
\end{itemize}
Since each meta-skill file uses the same Markdown representation as the task-skill files the pipeline already consumes, the same five agents that improve $s$ also improve $m$ when applied recursively.

A meta-skill is good insofar as it converts iterations into utility gains.  We make this precise through the \emph{meta-productivity} of $m$ at skill $s$, the expected per-child improvement over $K$ proposals,
\begin{equation}
  P(m \mid s) = \mathbb{E}\!\left[\tfrac{1}{K}\textstyle\sum_{k=1}^{K}\bigl(U(s'_k) - U(s)\bigr)\right],
\label{eq:productivity}
\end{equation}
estimated per node by the empirical mean $\hat{P}_v = \overline{\Delta U_{\text{children of }v}}$ (zero for nodes with no children). MetaSkill-Evolve jointly maximises task utility $U(s)$ and meta-productivity $P(m \mid s)$ across all branches: the fast loop improves the task skill $s$, and the slow loop improves the meta-skill $m$ that produces those improvements.

\begin{figure*}[t]
\centering
\includegraphics[width=\linewidth]{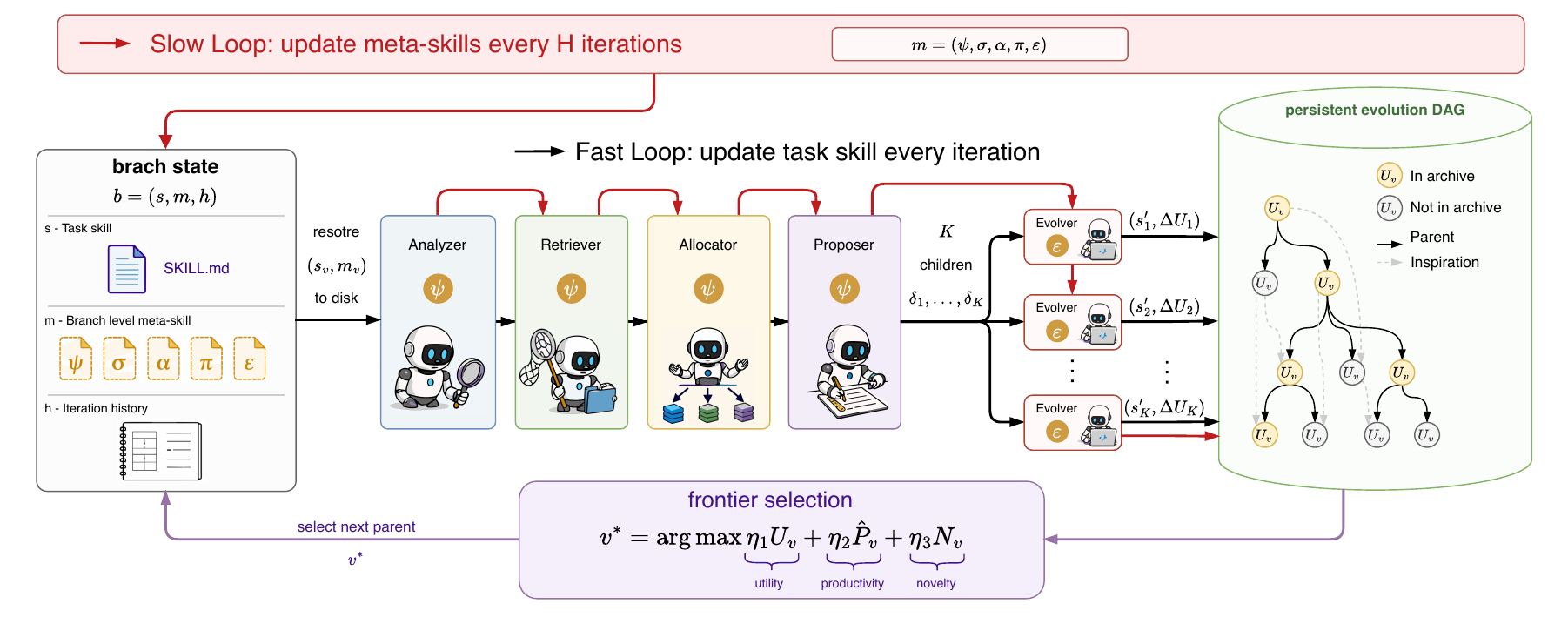}
\caption{System overview. The branch state $b=(s,m,h)$ (left) feeds the five-agent pipeline (centre), whose output is appended to the SQLite node graph (right). Frontier selection (Eq.~\ref{eq:select}) draws the next parent from the graph.}
\label{fig:system_overview}
\end{figure*}

\subsection{Evolution Graph and Frontier Selection}
\label{sec:graph}

We record the entire search history as a directed acyclic graph (DAG) $\mathcal{G}=(\mathcal{V},\mathcal{E})$ persisted in SQLite. Each node $v\in\mathcal{V}$ is one evaluated branch state and stores $(s_v, m_v, U_v, \Delta U_v, \phi_v)$ together with its branch path and selection counter. Edges are directed and of two kinds: a \emph{lineage} edge $u\!\rightarrow\!v$ marks $v$ as the child produced by evolving its parent $u$, and an \emph{inspiration} edge records a cross-branch node that $\sigma$ retrieved when proposing $v$. Both edge types point from an earlier node to a later one, so $\mathcal{G}$ is acyclic by construction: each node is created once, from already-existing nodes, and is never revised in place. Persisting $\mathcal{G}$ rather than keeping a fixed-size beam lets us revisit previously deprioritised lineages, supports cross-branch retrieval, and preserves full provenance for any final node.

A child enters the \emph{archive} (the pool of deployable states that also serves as the candidate set for future parents) only when it strictly improves on its parent, $\Delta U_v > 0$. Accuracy-neutral or regressing children ($\Delta U_v \leq 0$) are not eligible to be selected as parents, but are still persisted in $\mathcal{G}$: they preserve provenance and remain available to $\sigma$ as inspiration, so a neutral or failed edit can still inform a later proposal.

From the archive we refresh the frontier $\mathcal{F}$ each iteration to the top-$K$ nodes by
\begin{equation}
  v^* = \arg\max_{v \in \mathcal{F}}
    \bigl[\eta_1 U_v + \eta_2 \hat{P}_v + \eta_3 N_v\bigr],
\label{eq:select}
\end{equation}
where $\hat{P}_v$ is the meta-productivity estimate (Eq.~\ref{eq:productivity}) and $N_v = 1/(1 + \texttt{times\_selected}_v)$. Each term targets a distinct failure mode of greedy search:
\begin{itemize}[leftmargin=*,topsep=2pt,itemsep=1pt,parsep=0pt]
\item $U_v$ -- exploitation: prevents chasing volatile gain estimates on weak parents.
\item $\hat{P}_v$ -- trajectory quality: redirects effort from plateaued high points to nodes still generating useful descendants.
\item $N_v$ -- visitation cooling: a node selected $k$ times must outperform an unselected sibling by $\eta_3 k/(k{+}1)$ to be re-picked, preventing budget monopolies.
\end{itemize}
Setting any $\eta_i{=}0$ exposes the corresponding mode: the frontier locks on stagnated high-utility nodes ($\eta_2{=}0$), collapses to one lineage ($\eta_3{=}0$), or trusts noisy single-child gains as parent quality ($\eta_1{=}0$). Crucially, we do \emph{not} filter by lineage: diversity is a property of the score, not a structural constraint.

\subsection{Fast Timescale: Task-Skill Evolution}
\label{sec:fast}

The fast loop (Algorithm~\ref{alg:fast}) runs one task-skill iteration on a frontier parent $v$. Before invoking any agent, the runtime restores the selected branch's task and meta snapshots $(s_v, m_v)$ to disk.  Thus the SQLite DAG, rather than whatever files remain in the working tree, defines branch state: each branch starts from its recorded snapshot and is evaluated in isolation, preventing leakage between lineages. We then score $s_v$ on the training batch and take its \emph{worst-scoring} example as the diagnostic target: a deliberately high-signal choice, hedged against outliers because each resulting edit is judged by its validation gain $\Delta U_v$ rather than by that single training case. This example drives the five-agent pipeline:
\begin{itemize}[leftmargin=*,topsep=2pt,itemsep=1pt,parsep=0pt]
\item \textbf{Analyzer ($\psi$)}: emits a tag $\phi$ and a free-form analysis $a$; the tag vocabulary is itself maintained by $\psi$ and revised by the slow loop.
\item \textbf{Retriever ($\sigma$)}: ranks a $\phi$-matched candidate pool by tag similarity, over-fetching to $3{\times}$ the inspiration budget, then LLM-re-ranks this wider pool down to the inspirations $\mathcal{I}$ handed to the Proposer; the breadth/depth balance is itself a learned object.
\item \textbf{Allocator ($\alpha$)}: chooses $K \in [1, K_{\max}]$, widening search after stagnation ($\hat{P}\approx0$) and contracting after a productive edit.
\item \textbf{Proposer ($\pi$)}: for each of the $K$ children emits an edit $\delta$; when $K{>}1$ a diversity hint steers the $k$-th proposer toward a distinct intervention angle, reducing near-duplicate children.
\item \textbf{Evolver ($\varepsilon$)}: translates $\delta$ into file writes via \texttt{skill\_tools} and verifies the result with a before/after hash check that flags edits leaving the target files unchanged.
\end{itemize}
Each child $s'_k$ is evaluated on $\mathcal{D}_{\text{val}}$ to obtain its gain $\Delta U_k$. Every $H$ iterations the slow loop (\S\ref{sec:slow}) then refreshes the meta-skill $m_v$; the $K$ children are committed to $\mathcal{G}$ carrying this refreshed meta-skill, and the frontier is re-synchronised before the next iteration.

\begin{algorithm}[t]
\caption{Fast timescale: one task-skill iteration}
\label{alg:fast}
\small
\begin{algorithmic}[1]
\Require Frontier parent $v$; train batch $\mathcal{D}_\text{train}$, val batch $\mathcal{D}_\text{val}$
\Ensure $K$ committed child nodes (those with $\Delta U > 0$ enter the archive)
\Phase{Restore and evaluate}
\State Restore snapshots $s_v$ (task), $m_v$ (meta) to disk
\State $\mathcal{E} \leftarrow \text{Eval}(s_v, \mathcal{D}_\text{train})$ \Comment{collect failures}
\If{$\mathcal{E}$ has no failures} \Return \texttt{all\_passed} \EndIf
\State $f \leftarrow \arg\min_{e \in \mathcal{E}} \text{score}(e)$ \Comment{worst case}
\Phase{Diagnose and plan (3 agents)}
\State $\phi, a \leftarrow \text{Analyzer}(f,\, m_v.\psi)$ \Comment{tag, analysis}
\State $\mathcal{I} \leftarrow \text{Retriever}(\phi,\, b_v,\, m_v.\sigma)$ \Comment{inspiring nodes}
\State $K \leftarrow \text{Allocator}(h_v,\, a,\, \mathcal{I},\, m_v.\alpha)$ \Comment{child budget}
\Phase{Propose and evolve ($K$ children)}
\For{$k = 1 \ldots K$}
  \State Restore $s_v$
  \State $\delta \leftarrow \text{Proposer}(f,\, a,\, \mathcal{I},\, m_v.\pi)$
  \State $s'_k \leftarrow \text{Evolver}(s_v,\, \delta,\, m_v.\varepsilon)$
  \State $U'_k \leftarrow \text{Eval}(s'_k,\, \mathcal{D}_\text{val})$; \ \ $\Delta U_k \leftarrow U'_k - U_v$
\EndFor
\Phase{Interleave slow loop, commit and sync}
\If{$t \bmod H = 0$} $m_v \leftarrow$ update meta-skill (Alg.~\ref{alg:slow}) \EndIf
\State Commit children $\{\langle s'_k,\ \Delta U_k,\ m_v\rangle\}_{k=1}^{K}$ to $\mathcal{G}$
\State $\mathcal{F} \leftarrow \text{SyncFrontier}(\mathcal{F})$ \Comment{next parent drawn by Eq.~\ref{eq:select}}
\end{algorithmic}
\end{algorithm}

\subsection{Slow Timescale: Meta-Skill Evolution}
\label{sec:slow}

Updating $m$ at every iteration would expose the meta-skill to the same single-example noise that drives task-skill evolution. The slow loop (Algorithm~\ref{alg:slow}) instead fires once every $H$ fast iterations and aggregates over that horizon, trading reactivity for stability. Its driving signal is the empirical meta-productivity $\hat{P}(m \mid s) = \tfrac{1}{|\mathcal{H}|}\sum_{u \in \mathcal{H}} \Delta U_u$ over the last $H$ descendants $\mathcal{H}$. We fold $\hat{P}$ together with the tags, diagnoses, and outcomes of that window into a synthetic \emph{meta-failure trace} $f_m$: the improvement history reshaped to look like one failing training example, so that a single Analyzer prompt serves both timescales.

We then re-run the same five-agent pipeline on $f_m$, switching its \emph{target object} from task-skill files to the hidden meta-skill files. The Analyzer names the single most-implicated component of $\{\psi,\sigma,\alpha,\pi,\varepsilon\}$; this diagnosis fixes the failure tag $\phi_m$ that steers retrieval, but it does \emph{not} narrow the edit scope. The Retriever surfaces cross-branch lineages whose meta-failure tags match $\phi_m$, the Allocator sets the round budget $K_m$, and the Proposer and Evolver then co-edit \emph{all five} files per round. The resulting snapshot of all five files becomes the child's \texttt{meta\_state\_json}.  Each branch therefore carries its own lineage-local $m$, and the sole channel by which one lineage's improvement policy reaches another is this meta-level retrieval, so escape strategies propagate between lineages without any shared global state. Three details separate this from a plain fast-loop run:
\begin{itemize}[leftmargin=*,topsep=2pt,itemsep=1pt,parsep=0pt]
\item \textbf{Constrained Analyzer.} A null or task-skill diagnosis triggers a round-robin fallback over $\{\psi,\sigma,\alpha,\pi,\varepsilon\}$, so the slow loop never aborts on an unusable target nor silently degrades into a redundant fast-loop run.
\item \textbf{Whole-$m$ rewrite.} Each child edits \emph{all} five meta-skill files in one step (Proposer sequential, Evolvers parallel), preserving cross-component coherence; for instance, a $\pi$ edit assuming a finer tag vocabulary is co-applied with the matching $\psi$ edit.
\item \textbf{Accumulating children.} Child $k{+}1$ reads the files as written by child $k$, not by the parent; this moving target drives incremental refinement rather than $K_m$ independent overwrites that average back to the parent.
\end{itemize}

\begin{algorithm}[t]
\caption{Slow timescale: meta-skill update, at $t \bmod H = 0$}
\label{alg:slow}
\small
\begin{algorithmic}[1]
\Require Branch history $\mathcal{H}$ (last $H$ children); meta-skill $m$; parent $v$;
         names $\mathcal{M}{=}\{\psi,\sigma,\alpha,\pi,\varepsilon\}$
\Ensure Snapshot of all five meta-skill files
\Phase{Build meta-failure trace}
\State $\hat{P} \leftarrow \tfrac{1}{|\mathcal{H}|}\sum_{u \in \mathcal{H}} \Delta U_u$ \Comment{meta-productivity}
\State $f_m \leftarrow$ trace of tags, diagnoses, outcomes over $\mathcal{H}, \hat{P}$
\Phase{Diagnose and plan (target $\in \mathcal{M}$)}
\State $\phi_m, a_m \leftarrow \text{Analyzer}(f_m,\, m)$ \Comment{round-robin fallback}
\State $\mathcal{I}_m \leftarrow \text{Retriever}(\phi_m,\, b_v,\, m.\sigma)$
\State $K_m \leftarrow \text{Allocator}(\mathcal{H},\, a_m,\, \mathcal{I}_m,\, m.\alpha)$
\Phase{Whole-$m$ rewrite ($K_m$ accumulating children)}
\For{$k = 1 \ldots K_m$}
  \State Render meta-files from disk \Comment{reflects child $k{-}1$}
  \For{$j \in \mathcal{M}$} \Comment{Proposer: sequential}
    \State $\delta_m^{(j)} \leftarrow \text{Proposer}(f_m,\, a_m{\mid}\text{tgt}{=}j,\, \mathcal{I}_m,\, m.\pi)$
  \EndFor
  \State $\{m^{(j)}\} \leftarrow \text{ParallelEvolver}(\{\delta_m^{(j)}\}_{j \in \mathcal{M}},\, m.\varepsilon)$ \Comment{one worker/file}
\EndFor
\State \Return snapshot of $\{m^{(j)}\}_{j \in \mathcal{M}}$
\end{algorithmic}
\end{algorithm}

\section{Experiments}
\label{sec:experiments}

\subsection{Setup}
\label{sec:setup}

\paragraph{Benchmarks and backbone.}
We evaluate on three agentic benchmarks chosen to span complementary capabilities: \textbf{OfficeQA}~\citep{opsahlong2026officeqa}, \textbf{SealQA}~\citep{pham2026sealqaraisingbarreasoning}, and \textbf{ALFWorld}~\citep{shridhar2021alfworld}.

Within the evolution loop, each benchmark file is split by stratified sampling over its category column into three disjoint partitions: a training partition (failure mining), a validation partition (child scoring and best-skill selection), and a held-out test partition that the loop never observes. We then report accuracy of the selected skill on the held-out test partition through a separate \texttt{benchmark}-mode pass. 

A single frozen base model, Gemma-4 31B~\citep{google2026gemma431b}, serves all five pipeline agents (Analyzer, Retriever, Allocator, Proposer, Evolver); no agent is fine-tuned, so all gains are attributable to evolved skills and meta-skills.

\paragraph{Baselines.}
We compare four configurations, all sharing the same backbone:
\begin{itemize}[topsep=2pt,itemsep=1pt,leftmargin=*]
  \item \textbf{No-Skill}: the base agent with no skill loaded and
        reflection disabled. Quantifies the raw backbone.
  \item \textbf{Static Skill}: the same agent loaded with our
        hand-authored initial skill, held fixed for the entire run.
        Isolates the value of a skill artifact \emph{per se}.
  \item \textbf{Single-Level Evolution}: our fast loop with the slow
        loop frozen ($K_{\max}{=}1$, no cross-branch sharing, no
        meta-skill updates). Isolates the contribution of task-skill
        evolution alone.
  \item \textbf{MetaSkill-Evolve} (ours): the full two-timescale system.
\end{itemize}

\subsection{Main Results}
\label{sec:main}

\begin{figure}[t]
\centering
\includegraphics[width=\linewidth]{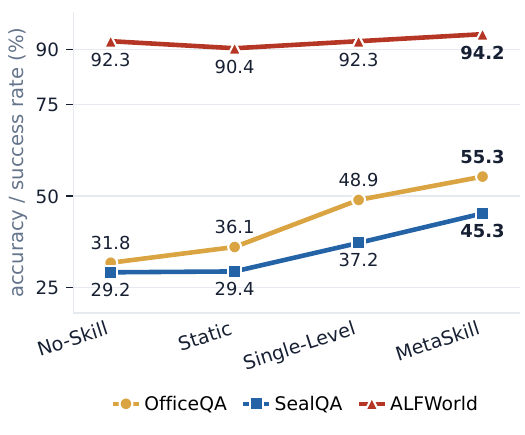}
\caption{\textbf{End-to-end held-out test accuracy on the three benchmarks.} All conditions share the Gemma-4 31B backbone; only the skill-evolution strategy differs. Red annotations mark MetaSkill-Evolve's gain over No-Skill; the No-Skill$\,\to\,$Ours ordering improves on every benchmark, with the largest margins on the two QA tasks where the backbone has the most headroom.}
\label{fig:main_results}
\end{figure}

Table~\ref{tab:main} and Fig.~\ref{fig:main_results} report held-out test accuracy across the four conditions on all three benchmarks. The two QA benchmarks, where the backbone has the most headroom, tell the cleanest story. \emph{First}, the static skill is worth several points over the raw backbone on OfficeQA (+4.31) and is roughly neutral on SealQA (+0.24). \emph{Second}, replacing the fixed skill by single-level evolution (our fast loop with the slow loop frozen) adds a further +12.85\,/\,+7.80 points, isolating the contribution of the inner evolution loop and the SQLite-backed evolution graph. \emph{Third}, switching on the slow loop lifts performance by another \textbf{+6.38\,/\,+8.05} points, a gain attributable to meta-skill adaptation, since the only difference from the Single-Level baseline is whether the meta-skill files $\{\psi,\sigma,\alpha,\pi,\varepsilon\}$ are themselves evolved. The progression No-Skill$\,\to\,$Static$\,\to\,$Single-Level$\,\to\,$MetaSkill-Evolve is monotonic on both QA benchmarks, so each design choice (adding a skill, evolving it, then evolving the procedure that evolves it) pays off independently: end-to-end, MetaSkill-Evolve improves held-out accuracy over the raw backbone by \textbf{+23.54} points on OfficeQA and \textbf{+16.09} on SealQA.

ALFWorld stresses the opposite regime: the backbone already solves most episodes (92.31\%), leaving little room to improve. The static skill slightly regresses ($-1.93$), and single-level evolution only recovers to the No-Skill baseline (92.31\%), so neither non-meta step yields a net gain over the raw backbone. The slow loop alone supplies the entire end-to-end improvement of \textbf{+1.92} points (to 94.23\%): small in absolute terms, but it indicates that meta-skill adaptation remains the operative ingredient even once task-skill evolution has saturated.

\begin{table}[t]
\centering
\small
\caption{\textbf{Main results.} Held-out test accuracy (\%) on OfficeQA, SealQA, and ALFWorld; the test partition is never seen during evolution (App.~\ref{app:leakage}). ALFWorld reports the aggregate task success rate. All rows share the Gemma-4 31B backbone; only the skill-evolution strategy differs.}
\label{tab:main}
\begingroup
\setlength{\tabcolsep}{6pt}
\renewcommand{\arraystretch}{1.18}
\newcommand{\best}[1]{\cellcolor{msGreenLight}\textbf{#1}}
\newcommand{\second}[1]{\cellcolor{msGoldLight}\underline{#1}}
\arrayrulecolor{msRule}
\begin{tabular}{@{}lccc@{}}
\toprule
\textbf{Method} & OfficeQA & SealQA & ALFWorld \\
\midrule
\multicolumn{4}{@{}l}{\textcolor{msBlue}{\emph{Non-evolutionary baselines}}} \\
\quad No-Skill                       & 31.78 & 29.17 & 92.31 \\
\quad Static skill                   & 36.09 & 29.41 & 90.38 \\
\midrule
\multicolumn{4}{@{}l}{\textcolor{msBlue}{\emph{Self evolution}}} \\
\quad Single-Level
  & \second{48.94} & \second{37.21} & \second{92.31} \\
\rowcolor{msBlueLight}
\quad\textbf{\textcolor{msBlue}{MetaSkill-Evolve}}
  & \best{55.32} & \best{45.26} & \best{94.23} \\
\midrule
\quad\textcolor{msBlue}{\emph{$\Delta$ vs.\ No-Skill}}
  & \textbf{+23.54} & \textbf{+16.09} & \textbf{+1.92} \\
\bottomrule
\end{tabular}
\arrayrulecolor{black}
\endgroup
\end{table}

\begin{figure}[t]
\centering
\includegraphics[width=\linewidth]{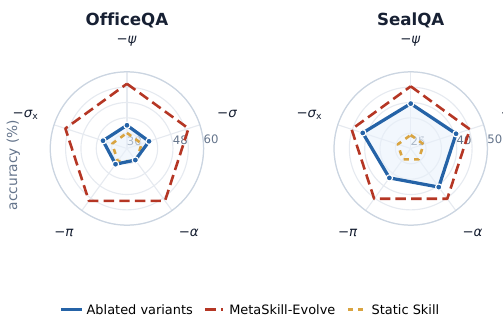}
\caption{\textbf{Component ablations on the two QA benchmarks.} Blue polygon: accuracy with one meta-skill component removed; dashed rings: full MetaSkill-Evolve and Static-Skill references. Dominant component differs by domain: $\alpha$ on OfficeQA, $\pi$ on SealQA.  ALFWorld in Table~\ref{tab:ablation}.}
\label{fig:component_ablation}
\end{figure}

\subsection{Component Ablations}
\label{sec:ablations}

To attribute the gains to specific design choices we disable one component at a time and re-run on all three benchmarks (Table~\ref{tab:ablation} in App.~\ref{app:ablations}; QA benchmarks visualised in Fig.~\ref{fig:component_ablation}). The configurations $-\psi$, $-\sigma$, $-\alpha$, $-\pi$ ablate the corresponding meta-skill component. In particular, $-\sigma$ removes the Retriever's inspiration policy entirely; the separate \emph{no cross-branch} condition (denoted $-\sigma_{\mathrm{x}}$ in Fig.~\ref{fig:component_ablation}) removes only cross-branch candidates while same-branch inspirations remain available. The \emph{no meta-updates} condition freezes the slow loop entirely. The Evolver $\varepsilon$ always executes (with $-\pi$ it consumes the raw analysis instead of a structured proposal), so $\varepsilon$ does not carry its own ablation row.

Fig.~\ref{fig:component_ablation} shows the component ablations on the two QA benchmarks. There are three findings. First, \emph{every typed component contributes}: no single-component ablation matches the full system, although which component matters most differs by domain. Second, on OfficeQA the \emph{allocation policy} $\alpha$ is the single most important component ($55.32{\to}35.58$, $-19.7$\,pts): the OfficeQA failure landscape contains pockets of related arithmetic errors where $\alpha$'s adaptive widening of the child budget after stagnation is what produces a successful child at all, and $\pi$ ($-17.7$\,pts) is a close second. Third, on SealQA the \emph{edit-proposal policy} $\pi$ is instead dominant ($45.26{\to}36.84$), where the gain hinges on the precise content of each edit rather than on how widely the search fans out.

\subsection{Meta-Update Horizon}
\label{sec:horizon}

The horizon $H$ governs the coupling between the two timescales: it sets how many fast task-skill iterations elapse between consecutive meta-skill evolutions (\S\ref{sec:slow}). The choice cuts two ways. Firing the slow loop too often exposes the meta-skill to the single-example noise that aggregating over $H$ is meant to filter; firing it too rarely lets the meta-skill go stale relative to the drifting task skill, so its broader rewrites land on edits the fast loop has already moved past and overwrite productive changes. To probe this trade-off while holding total compute fixed, we keep the number of meta-updates at three and scale the iteration budget with $H$, sweeping $H\in\{2,4,8\}$ (equivalently 6, 12, and 24 fast iterations). Fixing the meta-update count makes the $H{=}2$ point here (6 iterations, three meta-updates) a slightly different operating point from the five-iteration default (two meta-updates at $H{=}2$) behind Tables~\ref{tab:main} and~\ref{tab:ablation}: although all three are scored on the same held-out test partition, the extra iteration and meta-update mean its accuracies are not expected to coincide cell-for-cell with the full-system rows there.

\begin{figure}[t]
\centering
\includegraphics[width=\linewidth]{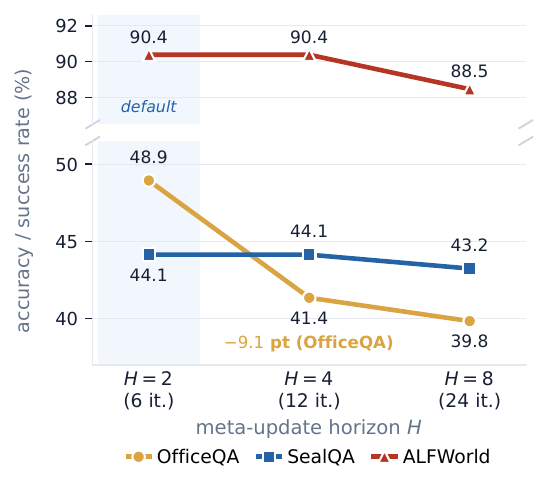}
\caption{\textbf{Meta-update horizon sweep.} Held-out test accuracy as the horizon $H$ (fast iterations between consecutive meta-skill evolutions) widens from 2 to 8, with the meta-update count held fixed at three, so total iterations $=3H$ (shown in parentheses). The broken $y$-axis separates near-ceiling ALFWorld (top) from the two QA accuracies (bottom). The default $H{=}2$ (shaded band) is best on every benchmark; OfficeQA is by far the most sensitive to a stale meta-skill.}
\label{fig:meta_horizon}
\end{figure}

Figure~\ref{fig:meta_horizon} (exact values in Table~\ref{tab:meta_horizon}, App.~\ref{app:hyperparams}) reports the sweep on the held-out test set. The tightest spacing $H{=}2$ is best on every benchmark, and accuracy falls as the meta-skill is refreshed less often, but the magnitude is strongly benchmark-dependent. OfficeQA is the most sensitive, shedding $9.1$ points from $H{=}2$ to $H{=}8$ ($48.94{\to}41.35{\to}39.84$); SealQA and ALFWorld are nearly flat between $H{=}2$ and $H{=}4$ and slip only at $H{=}8$ (a $0.9$- and $1.9$-point drop overall). All three horizons already aggregate over multiple iterations, escaping the per-iteration noise that motivates $H{>}1$; among them the most reactive schedule wins, which fixes our default at $H{=}2$.

\section{Conclusion}
\label{sec:conclusion}

We introduced MetaSkill-Evolve, a two-timescale framework in which every branch carries a task skill $s$ and a meta-skill $m=(\psi,\sigma,\alpha,\pi,\varepsilon)$ whose five components parameterise the improvement pipeline. Because each component of $m$ is itself a Markdown LLM-agent program, the same pipeline that rewrites $s$ on the fast loop refines $m$ on the slow loop, a bounded instance of recursive self-improvement that needs no extra model or training, while frontier selection $\eta_1 U_v + \eta_2 \hat{P}_v + \eta_3 N_v$ redirects search from plateaued branches to ones whose improvement policy is still productive. The result is a +23.54 / +16.09 / +1.92 point gain in held-out test accuracy over the raw Gemma-4 31B backbone on OfficeQA / SealQA / ALFWorld, of which the slow loop contributes +6.38 / +8.05 / +1.92, evidence that an agent's improvement policy admits the same search machinery as its task behaviour, and that separating \emph{what to do} from \emph{how to improve} keeps each loop's signal interpretable.

\section*{Limitations}
MetaSkill-Evolve is evaluated on three curated benchmarks; transfer to open-ended, long-horizon real-world tasks with noisier feedback is untested. The five-agent pipeline is itself fixed: we evolve the skills it produces but not its roles or wiring.  Meta-updates fire at a fixed horizon $H$.

\bibliography{refs}

\appendix

\section{Five-Agent Pipeline Details}
\label{app:pipeline}

All five agents are LLM-backed \texttt{ToolCallingAgent} instances sharing the same frozen base model; they load the skill catalog (task skills or meta-skills, depending on context) as additional context.

\paragraph{Analyzer ($\psi$).}
Given the execution trace and failure reason of the worst-scoring training example, the Analyzer produces (i) a structured root-cause analysis, (ii) a short failure tag $\phi$ (at most 15 words), and (iii) the target skill file to edit. It uses a three-layer recovery chain: primary parse of \texttt{final\_answer}, step-level output scanning across all agent steps, and a repair call with constrained \texttt{response\_format}. The \texttt{target\_skill} field is auto-derived from \texttt{relevant\_sections} when not explicitly provided, improving robustness with smaller models.

\paragraph{Retriever ($\sigma$).}
Given $\phi$ and the current branch path, the Retriever fetches same-branch and cross-branch candidates from the SQLite graph store, over-fetched to $3L_{\text{same}}$ and $3L_{\text{cross}}$ by tag similarity, and LLM-re-ranks them by relevance to the present failure, returning the subset it judges relevant.

\paragraph{Allocator ($\alpha$).}
Given recent $\Delta U$ values along the branch and the analysis, the Allocator chooses a child budget $K \in [1, K_{\max}]$, allocating more search effort on stagnation and less when recent edits have been productive.

\paragraph{Proposer ($\pi$).}
Given the failure analysis, the used skill materials, and the inspiring nodes, the Proposer produces a concrete edit proposal: target section, change, and rationale. When $K{>}1$, a diversity hint instructs it to take a distinct intervention angle from prior child proposals.

\paragraph{Evolver ($\varepsilon$).}
The Evolver reads the current skill file, writes the edited version, and verifies that the mutations are consistent with the proposal summary. After each apply, the skill registry is refreshed so subsequent evaluations use the updated skill.

\section{Meta-Skill Representation and Skill-Catalog Disclosure}
\label{app:representation}

Each meta-skill component $\psi, \sigma, \alpha, \pi, \varepsilon$ is stored as a \texttt{SKILL.md} file under the project skills directory (e.g., \texttt{skills/meta-analyzer/SKILL.md}) and snapshotted into the SQLite node graph alongside the task-skill snapshot, so each node carries a complete, self-contained record of both task-level and meta-level state at creation.  When a branch is selected for expansion, its meta-skill snapshot is restored from the node record before the five-agent pipeline runs, ensuring each branch applies its own lineage-specific improvement policy. Branches can thus diverge in their meta-level heuristics: one may have learned an aggressive edit policy for table-extraction failures while another developed conservative, incremental edits for arithmetic reasoning failures.

\paragraph{Skill-catalog progressive disclosure.}
Agents receive a compact catalog first (skill names and one-line summaries), then load the full \texttt{SKILL.md} only for the skill they identify as relevant, and load resource files on demand. This keeps context length manageable while preserving full expressiveness.

\section{Meta-Skill Prompt Templates}
\label{app:prompts}

Each meta-skill component is initialized as a brief Markdown document that describes its role in the evolutionary process.  For example, the initial diagnosis policy $\psi$ instructs the Analyzer to: (i) identify the primary failure class from the execution trace, (ii) distinguish between skill-addressable failures and base-model capability limits, and (iii) assign a short, specific failure tag.

\section{Hyperparameter Sensitivity}
\label{app:hyperparams}

\paragraph{Defaults.}
Frontier weights $\eta_1{=}1.0$, $\eta_2{=}0.5$, $\eta_3{=}0.25$; meta-update horizon $H{=}2$; iteration budget 5 fast iterations (two meta-updates at $H{=}2$); child budget $K \in [1, K_{\max}]$ with $K_{\max}{=}3$ and an initial $K{=}2$ that the Allocator adapts per step (this initial value is what the $-\alpha$ ablation freezes to); frontier size $K_{\mathcal{F}}{=}3$, with early stopping after 5 iterations without a frontier improvement. Category-aware round-robin training draws 6 categories per batch and 3 samples per category. Cross-branch sharing uses retrieval probability $p_{\text{cross}}{=}0.2$ and same-/cross-branch inspiration limits $L_{\text{same}}{=}3$, $L_{\text{cross}}{=}2$ (over-fetched at $3\times$ before LLM re-ranking, \S\ref{sec:fast}). Evaluation concurrency defaults to 16, raised to 128 for the QA benchmarks under the vLLM backend; it affects only throughput, not accuracy.


\paragraph{Sensitivity to cross-branch sharing.}
Sweeping the Retriever's cross-branch candidate limit confirms that moderate cross-branch retrieval consistently improves performance on both OfficeQA and SealQA; very aggressive cross-branch retrieval introduces noise by supplying irrelevant inspirations from branches working on structurally different failures.

\begin{table}[h]
\centering
\caption{Meta-update horizon sweep (held-out test accuracy, \%), the exact values plotted in Fig.~\ref{fig:meta_horizon}. $H$ is the number of fast task-skill iterations between consecutive meta-skill evolutions; the number of meta-updates is held fixed at three, so the iteration budget scales as $3H$. The tightest spacing $H{=}2$ is best on every benchmark; wider gaps let stale meta-rewrites overwrite productive edits.}
\label{tab:meta_horizon}
\small
\setlength{\tabcolsep}{6pt}
\begin{tabular}{lccc}
\toprule
\textbf{Benchmark} & \textbf{$H{=}2$} & \textbf{$H{=}4$} & \textbf{$H{=}8$} \\
                   & {\footnotesize(6 it.)} & {\footnotesize(12 it.)} & {\footnotesize(24 it.)} \\
\midrule
OfficeQA & \textbf{48.94} & 41.35 & 39.84 \\
SealQA   & \textbf{44.14} & 44.14 & 43.24 \\
ALFWorld & \textbf{90.38} & 90.38 & 88.46 \\
\bottomrule
\end{tabular}
\end{table}

\section{Full Ablation Tables}
\label{app:ablations}

Table~\ref{tab:ablation} expands the component-ablation summary of
\S\ref{sec:ablations}.

\begin{table}[h]
\centering
\caption{Component ablation across the three benchmarks (held-out test accuracy, \%), scored on the same partition as Table~\ref{tab:main}. The Evolver ($\varepsilon$) always executes; with $-\pi$ it consumes the raw analysis instead of a structured proposal. $-\sigma$ removes the Retriever's inspiration policy entirely; \emph{no cross-branch} removes only cross-branch retrieval candidates while same-branch candidates remain available. \emph{No meta-updates} freezes the slow loop, exactly reproducing the Single-Level row of Table~\ref{tab:main}.}
\label{tab:ablation}
\small
\setlength{\tabcolsep}{4pt}
\resizebox{\linewidth}{!}{%
\begin{tabular}{lccc}
\toprule
\textbf{Configuration} & \textbf{OfficeQA} & \textbf{SealQA} & \textbf{ALFWorld} \\
\midrule
Full MetaSkill-Evolve                             & \textbf{55.32} & \textbf{45.26} & \textbf{94.23} \\
$-\psi$ (\texttt{disable\_psi})                   & 39.09 & 39.63 & 88.46 \\
$-\sigma$ (no inspirations)                       & 39.09 & 40.54 & 88.46 \\
$-\alpha$ (\texttt{disable\_alpha}, $K{=}2$)      & 35.58 & 40.54 & 88.46 \\
$-\pi$ (\texttt{disable\_pi})                     & 37.59 & 36.84 & 86.54 \\
No cross-branch retrieval                         & 39.84 & 41.44 & 92.31 \\
No meta-updates (\texttt{disable\_meta\_updates}) & 48.94 & 37.21 & 92.31 \\
\bottomrule
\end{tabular}%
}
\end{table}

On ALFWorld the \emph{edit-proposal policy} $\pi$ is again the dominant component ($94.23\to86.54$), echoing SealQA: the gain hinges on the precise content of each edit rather than on how widely the search fans out. Cross-branch sharing carries the remainder of the meta-gain. Removing only cross-branch retrieval (\emph{no cross-branch}, $92.31$) returns accuracy exactly to the Single-Level baseline ($92.31$), so cross-branch transfer of reusable sub-routines (e.g.\ object-disambiguation and deferred-placement sub-skills) accounts for the whole $+1.92$ improvement there. As on the QA benchmarks, freezing the slow loop (\emph{no meta-updates}, $92.31$) reproduces the Single-Level row of Table~\ref{tab:main} exactly.

\section{Evaluation Protocol: Native Held-Out Test}
\label{app:leakage}

The accuracy figures in Table~\ref{tab:main} are computed on a held-out \emph{test} partition that the evolution loop never observes, so they measure generalisation to unseen rows rather than recall of evolved ones.

Before the loop runs, each benchmark file is stratified by its category column into three disjoint partitions: train (failure mining), val (child scoring and frontier selection), and test (the per-category remainder).

\section{Train/Validation Split Ratio Sensitivity}
\label{app:split_ratio}

We study how the loop's train/validation split affects the quality of the evolved task skill. All numbers in this section are recomputed on a \emph{common} held-out test set per benchmark: because the stratified split is seeded, a larger-holdout run's test set is a subset of a smaller one's, so we intersect the test items across the swept runs and re-score every configuration on that shared subset (no re-running). These subsets are smaller than the benchmark sets used in the main results, so absolute numbers differ slightly.

\paragraph{Training-ratio sweep.}
Table~\ref{tab:train_ratio} sweeps the training fraction with validation fixed at $0.10$ (dual-evolve, $5$ outer iterations). More training data yields only modest gains: ALFWorld rises monotonically ($87.7\to88.7\to90.6$), while SealQA and OfficeQA are essentially flat within noise ($\pm 1\sim4$ items on $n\approx87\sim107$).

\begin{table}[h]
\centering
\caption{Training-ratio sweep (val\,$=0.10$), accuracy (\%) on the common held-out test subset.}
\label{tab:train_ratio}
\small
\setlength{\tabcolsep}{6pt}
\begin{tabular}{lcccc}
\toprule
\textbf{Benchmark} & \textbf{tr\,=\,0.05} & \textbf{tr\,=\,0.10} & \textbf{tr\,=\,0.15} \\
\midrule
SealQA   & 45.98 & 43.68 & \textbf{48.28} \\
OfficeQA & 44.86 & 42.99 & \textbf{48.25} \\
ALFWorld & 87.74 & 88.68 & \textbf{90.57} \\
\bottomrule
\end{tabular}
\end{table}

\paragraph{Validation-ratio sweep on ALFWorld.}
Table~\ref{tab:val_ratio} enlarges the validation split ($0.10{\to}0.25$) at each training ratio. A larger validation set consistently \emph{improves} the evolved skill, by $+2.0$, $+4.3$, and $+2.5$ points at $\text{tr}{=}0.05/0.10/0.15$ respectively. This effect is specific to high-baseline benchmarks such as ALFWorld, where the base agent already solves most episodes out of the box: a small validation set then contains too few \emph{failures} to drive evolution. The analyzer sees almost no failing trajectories, so the diagnosis$\to$proposal$\to$selection cycle has essentially no signal to act on and the slow loop degenerates toward a no-op. Enlarging the validation split surfaces enough failure cases to make skill proposal and frontier selection informative, which in turn yields a measurably better skill on the held-out test set. The practical implication is that on saturated benchmarks the validation split must be large enough to expose failures; otherwise the evolution loop has nothing to learn from and cannot improve.

\begin{table}[h]
\centering
\caption{ALFWorld validation-ratio sweep, accuracy (\%) on the common held-out test subset. Enlarging the validation split exposes more failure cases and produces a better evolved skill.}
\label{tab:val_ratio}
\small
\setlength{\tabcolsep}{6pt}
\begin{tabular}{lcc}
\toprule
\textbf{train ratio} & \textbf{val\,=\,0.10} & \textbf{val\,=\,0.25} \\
\midrule
tr\,=\,0.05 & 87.74 & \textbf{89.69} \\
tr\,=\,0.10 & 88.68 & \textbf{92.96} \\
tr\,=\,0.15 & 90.57 & \textbf{93.02} \\
\bottomrule
\end{tabular}
\end{table}

\end{document}